\documentclass[letterpaper, 10 pt, conference]{ieeeconf}  
\usepackage[usenames]{color}
\usepackage{epsfig}
\usepackage{graphics}
\usepackage{amsmath}
\usepackage{cite}
\usepackage{amssymb}
\usepackage{multirow}
\usepackage{cite}
\usepackage{array}
\usepackage{pslatex} 
\usepackage{url}
\usepackage{caption}
\usepackage{lineno}
\usepackage{graphicx}  
\usepackage{setspace}
\usepackage{tikz}
\usepackage{hhline}
\usepackage{braket}
\usepackage{float}
\usepackage{hyperref}

\usepackage{siunitx}
\usepackage{booktabs}

\usepackage{tikz}
\usetikzlibrary{bayesnet}

\usepackage{mathtools}

\IEEEoverridecommandlockouts                              
\usepackage[utf8]{inputenc}
\usepackage[T1]{fontenc}

\newcommand\copyrighttext{%
  \footnotesize \textcopyright 2021 IEEE. Personal use of this material is permitted. Permission from IEEE must be obtained for all other uses, in any current or future media, including reprinting/republishing this material for advertising or promotional purposes, creating new collective works, for resale or redistribution to servers or lists, or reuse of any copyrighted component of this work in other works.}  
\newcommand\copyrightnotice{%
\begin{tikzpicture}[remember picture,overlay]
\node[anchor=south,yshift=10pt] at (current page.south) {\fbox{\parbox{\dimexpr\textwidth-\fboxsep-\fboxrule\relax}{\copyrighttext}}};
\end{tikzpicture}%
}

\title{\LARGE \bf
Latent Factor Decomposition Model: Applications for Questionnaire Data*
}

\author{Connor J. McLaughlin$^{1}$, Efi G. Kokkotou$^{2}$, Jean A. King$^{3}$,  Lisa A. Conboy$^{4}$, Ali Yousefi$^{1}$ 
\thanks{*This work was supported in part by the NIH R01 Grant, AT001414-01, ``Enhancing the Placebo Effect in Irritable Bowel Syndrome''.}
\thanks{$^{1}$Connor J. McLaughlin (cjmclaughlin@wpi.edu) and Ali Yousefi are with the Department of Computer Science at Worcester Polytechnic Institute.}
\thanks{$^{2}$Efi G. Kokkotou is with the Beth Israel Deaconess Medical Center.}
\thanks{$^{3}$Jean A. King is with the Department of Biology and Biotechnology at Worcester Polytechnic Institute.}
\thanks{$^{4}$Lisa A. Conboy is with the Massachusetts College of Pharmacy and Health Services.}
}

\begin{document}

\maketitle
\copyrightnotice
\thispagestyle{empty}
\pagestyle{empty}

\begin{abstract}

The analysis of clinical questionnaire data comes with many inherent challenges. These challenges include the handling of data with missing fields, as well as the overall interpretation of a dataset with many fields of different scales and forms. While numerous methods have been developed to address these challenges, they are often not robust, statistically sound, or easily interpretable. Here, we propose a latent factor modeling framework that extends the principal component analysis for both categorical and quantitative data with missing elements. The model simultaneously provides the principal components (basis) and each patients' projections on these bases in a latent space. We show an application of our modeling framework through Irritable Bowel Syndrome (IBS) symptoms, where we find correlations between these projections and other standardized patient symptom scales. This latent factor model can be easily applied to different clinical questionnaire datasets for clustering analysis and interpretable inference.

\end{abstract}

\section{INTRODUCTION}

This application proposes a solution to issues faced in clinical datasets built from patient questionnaires. The first issue is in the prominence of missing data, especially in longer questionnaires. Additionally, these questions often result in a dataset with many fields of both categorical and continuous data, which renders them difficult to interpret. The question of how best to combine individual fields to summarize the state of a patient's condition is the focus of this research. The efficacy of our method is demonstrated through a dataset on Irritable Bowel Syndrome (IBS) \cite{Lembo2009}.

We can approach inference of questionnaire data as a problem of dimensionality reduction, i.e. how best to transform the many clinical measurements into fewer, representative variables. One common method for this transformation is to take a summation of patient responses. These aggregate measures can be effective for expressing the magnitude of patient condition severity, as done in the Whorwell Severity Scale for IBS \cite{Whorwell}. This method, however, concedes information about the specific condition of each patient; two patients with dissimilar symptoms may be summarized with the same aggregate score. To capture this underlying structure, we must utilize more robust dimensionality reduction techniques. 

The most prominent method for dimensionality reduction is  Principal Component Analysis (PCA) \cite{PCA}, which reduces a dataset to a set of uncorrelated variables which explain the most variance present in the data. While this technique is useful for dimensionality reduction without losing information, this approach is not equipped to handle categorical data, as found in questionnaires. Multiple Correspondence Analysis (MCA) \cite{MCA} is equivalent to PCA for nominal data, and may also be extended to handle both categorical and continuous measures. MCA may also be extended to handle mixed data types (combinations of categorical and continuous measures), known as Factor Analysis of Mixed Data (FAMD). The drawback to these methods is that the handling of missing data is not embedded in the method and it requires a secondary process. Furthermore, these methods are developed with pre-set assumptions on the categorical and quantitative data distributions, which may not hold for certain datasets. The outcome of these measures is a point estimate of data points' coordinates in the principal component space, which can mask the statistical significance of these projections. More recently, flexible techniques such as t-SNE \cite{TSNE}, UMAP \cite{umap}, or VAEs \cite{hi-vae} may also be used to reduce the dimensionality of mixed data, but the interpretability of the learned transformation is complex. 

To answer challenges in the analysis of questionnaire data, we propose a probabilistic model which may simultaneously handle both missing data and mixed data types in the search of principle components and new data points' coordinates. This model reduces questionnaire responses into a low dimensional approximately-orthogonal basis (principal components), which allows for the separation of different underlying conditions. This mapping allows clinicians to better capture the condition affecting each patient. Our principle components and corresponding point coordinates provide interpretable pictures of how different symptoms are grouped, and how much of each symptom group is present for each patient. Furthermore, the latent variables for each patient may be used as a predictor of treatment outcome, or as an alternative measure for monitoring disease progression. 

The structure of our paper is as follows: Section \ref{section:methods} introduces our modeling framework formulation. Section \ref{section:results} shows the interpretation and results of our model for IBS. Section \ref{section:conclusion} provides an analysis of the significance of our model.

\section{MATERIALS AND METHODS}
\label{section:methods}

\subsection{Proposed Model Framework}
Our method finds the first $d$ principal components in patient questionnaire data with $n$ patients and $m$ symptoms. Vector $\mathbf{y_n}$ has $m$ elements representing the $n^{th}$ patient's responses to the questionnaire. Each element of the vector represents a symptom, which may be present, not present, or not reported. Vector $\mathbf{x_n}$ with $d$ elements, is the latent weight vector for the $n^{th}$ patient, where the principal components are defined by a matrix $\mathbf{A_{mxn}}$ and intercept vector $\mathbf{b_{mx1}}$. Using $\mathbf{A}$ and $\mathbf{x_n}$, the probability of observing a $\mathbf{y_n}$ is defined by: 

\begin{subequations}
\label{eq:optim}
\begin{align}
    P(y_{nj} | x_{n}) \sim \text{Bernoulli}(P_{nj})_{j=1...m} \\ 
        \text{logit} \  P_{nj} = A_{j}x_{n} + b_{j}
\end{align}
\end{subequations}

Where $y_{nj}$ is the $j^{th}$ element of the vector $\mathbf{y_n}$. $\mathbf{A_j}$ is the $j^{th}$ row of the basis $\mathbf{A}$, and vector $\mathbf{b}$ is the expected outcome of the $j^{th}$ symptom, in logit scale, across all patients. Note that $\mathbf{x_n}$ is not observed, and it is a latent variable in our model.

Here, we assume the latent variables follow a multivariate normal with common prior $\mu_0$ and covariance of $\Sigma_0$: $x_n \sim N(\mu_0, \Sigma_0)$. 

The likelihood of observing the symptoms $\mathbf{Y_{nj}}$ given $\mathbf{x_n}$ is defined by:

\begin{equation}
L(y_{nj} | x_n) =
 \begin{cases}
      P_{nj} & \text{if $Y_{nj} = 1$ (present)} \\
      1 - P_{nj} & \text{if $Y_{nj} = 0$ (not present)}\\
      1 & \text{not reported}
    \end{cases}
\label{Eq:Likelihood_Symptoms}
\end{equation}

and the full likelihood of observed symptoms across patients is defined by:

\begin{equation}
    L(y_{1...N}; x_{1..N}, \mu_0, \Sigma_0, A, b) = \prod_{n=1}^{N}\prod_{j=1}^{M} L(y_{nj} | x_n)
\end{equation}

We assume the symptoms are independent given $\mathbf{x_n}$, and that each participants responses are independent from others. The posterior distribution on our model parameters and our latent variable is defined by:

\begin{equation}
\begin{multlined}
    P(x_{1...N}, A, b, y_{1...N} ; \mu_0, \Sigma_0) 
    \\
    \propto L(y_{1...N}; x_{1...N}, \mu_0, \Sigma_0, A, b) \prod_{n=1}^{N} P(x_n ; \mu_0, \Sigma_0)
\end{multlined}
\label{eq:4}
\end{equation}

where $P(x_n ; \mu_0, \Sigma_0)$ is the likelihood of $\mathbf{x_n}$ given ($\mu_0, \Sigma_0$) using the multivariate normal prior. 

We can also include continuous measurements in our model. Let vector $\mathbf{z_n}$, with $k$ elements, represent the $n^{th}$ patient's continuous measurements. Each element, as with our categorical data, may be measured or not reported. The likelihood of observing measurement $\mathbf{z_{nk}}$ given $\mathbf{x_n}$ is defined by:

\begin{equation}
L(z_{nk} | x_n) =
 \begin{cases}
      \frac{1}{\sqrt{2\pi}\sigma_k}\exp
      {\frac
      {-(z_{nk}-A_{k}x_{n}-b_{k})^{2}}
      {2\sigma_{k}^{2}}}
      & \text{if measured} \\
      1 & \text{not reported}
    \end{cases}
\label{eq:5}
\end{equation}

where we assume $\mathbf{A}$ and $\mathbf{b}$ to have $m+k$ rows. We assume $\mathbf{z_{nk}}$ follows a normal distribution with mean $\mathbf{A_k}\mathbf{x_n} + \mathbf{b_k}$ and unknown variance $\sigma_k$. Other continuous distributions, such as log-normal or exponential, could also be used to model $\mathbf{z_{nk}}$, depending on the type of measurements. For cases of mixed data, our model likelihood is defined by:

\begin{equation}
\begin{multlined}
    L(z_{1...N}, y_{1...N}; x_{1...N}, \mu_0, \Sigma_0, A, b) = L(y_{1...N}; x_{1...N}, \mu_0, \Sigma_0, A, b) \\
     * \prod_{n=1}^{N}\prod_{k=1}^{K} L(z_{nk} | x_n) \;\;\;\;
\end{multlined}
\label{eq:6}
\end{equation}

where we assume categorical and continuous measurements are statistically independent given $\mathbf{x_n}$. We can similarly define the posterior of the model parameters like in equation \ref{eq:4}, where we only have categorical data. Note that we now have a new set of parameters $\sigma_{k}^{2}$ which need to be estimated along other parameters in the model.

\begin{figure}[h]
\begin{center}
\begin{tikzpicture}
  \node[latent]                               (X) {$X$};
  \node[obs, below=of X, xshift=-1.5cm]                               (Y) {$Y$};
  \node[obs, xshift=-1.5cm]                               (Z) {$Z$};
  \node[latent, above=of X] (T) {$\mathbf{\theta}$};
  \node[latent, xshift=-2cm, yshift=0.75cm, below=of Z] (O) {$A, b$};

  \edge {X} {Z} ; %
  \edge {X} {Y} ; %
  \edge {T} {X} ; %
  \edge {O} {Z} ;
  \edge {O} {Y} ;

  \plate {xyz} {(X)(Z)(Y)} {$N$} ;

\end{tikzpicture}

\caption{Graphical model for latent factor model. $\theta$ is the set of parameters characterizing prior distributions of the latent variable \textbf{X}; \textbf{Z} and \textbf{Y} are continuous and categorical observed variables. \textbf{A} and \textbf{b} are the observation model free parameters.}
\label{graphical-model}
\end{center}
\end{figure}
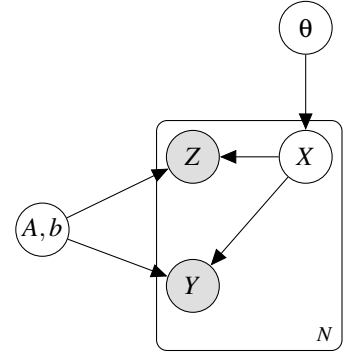

The latent factor model defined in equation \ref{eq:6} can represented as a graphical model as shown in Figure \ref{graphical-model}. Note that $\theta = \{\mu_0, \Sigma_0\}$ is the prior on our latent variable $\mathbf{X}$.

\vspace{-0.75cm}
\subsection{Parameter Optimization}
Our objective is to find $\mathbf{A}$ and $\mathbf{b}$ along with $\mathbf{x_{1...N}}$ which maximize the likelihood defined in equation \ref{eq:4} or \ref{eq:6}. This objective can be expanded to mixed data by including $\sigma_{1...k}$ in the optimization step. The optimization function is defined by:

\begin{equation}
\begin{multlined}
    \underset{x_{1...N}, A, b}{\mathrm{argmax}} \  P(x_n, A, b, y_n; \sigma_0, \Sigma_0) \\
    s.t. ||A^{T}A - I||_{F}^{2} < \gamma
\end{multlined}    
\end{equation}

where $||A^{T}A - I||_{F}^{2}$ is the Frobenius norm and $\gamma$ is a small positive number. This norm will imply matrix $\mathbf{A}$ to be approximately orthonormal.

When $\mathbf{x_n}$ are known, we can use out-of-box numerical methods to find $\mathbf{A}$ and $\mathbf{b}$ elements. Here, we must simultaneously estimate $\mathbf{x_{1..N}}$ and $\mathbf{A}$ and $\mathbf{b}$ elements; thus we use expectation-maximization (EM) techniques to recursively find posterior estimation of $\mathbf{x_{1..N}}$ and $\mathbf{A}$ and $\mathbf{b}$ elements that maximize the posterior, as in \cite{EM-Algorithm-Missing-Data} \cite{em-algorithm}. The EM technique is an established solution for maximum likelihood estimation in the presence of latent and/or missing data points. We use sampling techniques, e.g. Monte-Carlo, to find the posterior of $\mathbf{x_{1..N}}$ and use those samples to find what $\mathbf{A}$ and $\mathbf{b}$ will maximize the expected log of the posterior. Our implementation can be found at the github link: \url{https://github.com/YousefiLab/LatentFactorQuestionnaire}.

In the following section, we show an application of our methodology in the IBS dataset. We focus on reported IBS symptom data, so our model contains only categorical data.

\section{RESULTS}
\label{section:results}
\subsection{IBS Dataset}
We studied the dataset from the clinical trial initially described in \cite{Lembo2009}. This includes samples from 230 IBS patients, with measurements taken at three time points throughout the study. Measurements consist primarily of questionnaire data based on IBS pain and reported symptoms. The primary outcome measures analyzed in previous studies are the IBS Global Improvement Scale \cite{ibsglobalimprovement}, IBS Symptom Severity Scale \cite{Whorwell}, and the IBS Quality of Life (IBS-QOL) measure \cite{QOL}. Each of these scales are formed through questionnaire data, either in single questions or through the aggregation of many questions (e.g. 34 questions for IBS-QOL). In our analysis, we sought to create a more robust representation of patients' overall symptoms, based on a series of 19 binary symptom variables. We focus our analysis on the group of patients who received treatment (non-waitlist group), in the form of sham acupuncture and patient-practitioner interaction. We excluded data points where the patient dropped out of the study after the baseline measurements (n=126). Our proposed model is especially useful for the analysis of conditions such as IBS, which have high inter-patient variability \cite{ibsvariability}. 

\vspace{-0.2cm}
\subsection{Basis and Latent Variables}
\begin{figure}[h]
    \begin{center}
    \includegraphics[width=\columnwidth]{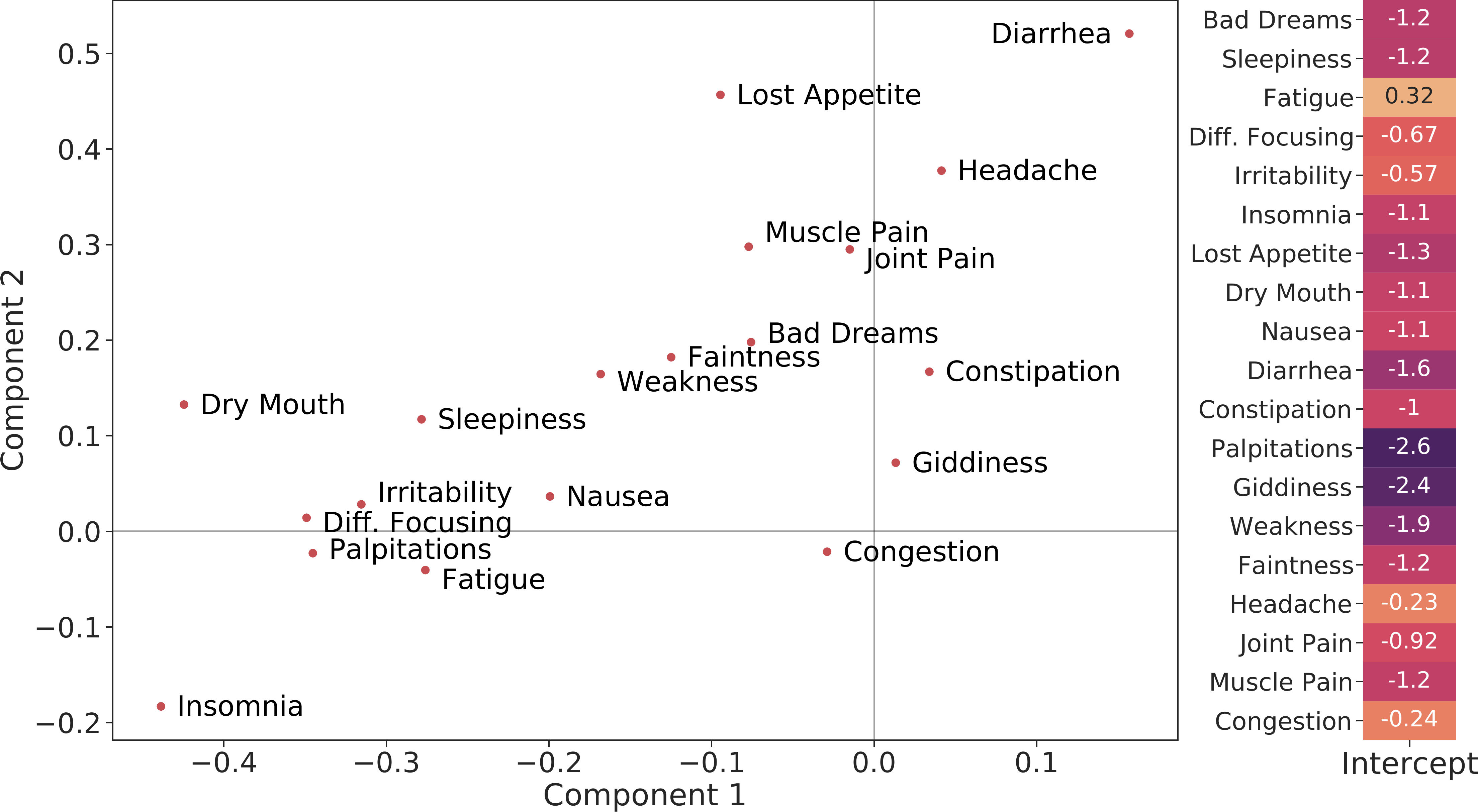}
    \caption{Fitted parameters $\mathbf{A}$, $\mathbf{b}$ on IBS symptom data. (Left) Basis $\mathbf{A}$ shown with each component in logit space, meaning symptoms with higher values in basis $n$ have higher probability given positive $x_n$. (Right) The intercept term $\mathbf{b}$, which represents the expected log-odds of each symptom across patients.}
    \label{fig:IBS-parameters}
    \end{center}
    \vspace{-0.6cm}
\end{figure}

We applied our modeling framework on the IBS dataset under the assumption that two principal components may capture the structure of reported symptoms. Figure \ref{fig:IBS-parameters} shows each of the questionnaire symptoms plotted in the transformed principle components coordinates. We see clear groupings of symptoms along each of these components. Along the first component, we note the importance of symptoms associated with the lack of psychological or sleep-related issues, such as insomnia, dry mouth, and difficulty concentrating (lower left of figure). Along the second component, we note the importance of symptoms generally associated with physical issues, such as diarrhea, headache, muscle/joint pain, and loss of appetite (upper right of figure). This offers a clear interpretation for physicians, as we can now understand how each latent component describes the patients' overall disease burden. 

Our centering term $\mathbf{b}$ provides the expected log-odds of each symptom across all patients. In figure \ref{fig:IBS-parameters} we can see that headache has the largest intercept term, indicating it may be a more common symptom across patients, whereas palpitations has a strong negative log-odds, indicating that it may be a rarer and more severe symptom. 

Given this context, latent variable estimates for each patient provide insight into the observed symptoms for each patient. For example, a patient with a higher value of $\mathbf{x_1}$ and a lower value of $\mathbf{x_2}$ may have more physical pain symptoms and few psychological or sleep symptoms. These variables provide effective summary measures which may result in distinctive clusters, showing commonalities across patient groups.

In the next section, we discuss how these latent variables relate to other known measures of IBS. 

\subsection{Correlation at Baseline}
We subsequently examined the potential correlation of our summary statistics ($\mathbf{x_1}$ and $\mathbf{x_2}$) with other patient metrics measured at baseline using established scales. In table \ref{table:correlation}, we show Pearson correlation coefficients of $\mathbf{x_1}$ and $\mathbf{x_2}$ with IBS-specific and non-specific measures of physical pain and psychological distress. Recall that $\mathbf{x_1}$ indicates a lack of psychological symptoms, and $\mathbf{x_2}$ indicates the presence of physical pain symptoms. We note that there is a consistent negative correlation with $\mathbf{x_1}$ across all metrics, suggesting perhaps that alleviation of the psychological cluster of symptoms might have a positive impact on the overall patient symptomatology. For $\mathbf{x_2}$, there is a positive correlation with anxiety and both general and IBS-specific pain, which indicates that treatment of diarrhea, lost appetite, and headache may result in reduced anxiety and pain. 

In a broader sense, these correlation results imply that $\mathbf{x_1}$ and $\mathbf{x_2}$ bring new and meaningful summary metrics for each patient which can be used for finding associations with symptom clusters and established outcome measures.

\begin{table}[ht]
    \sisetup{table-format = -1.3, table-space-text-post=$^{*}$}
	\begin{center}	
		\resizebox{1\columnwidth}{!}{
			\begin{tabular}{m{0.6\columnwidth} m{0.2\columnwidth} m{0.2\columnwidth} @{}l
                 S[table-format = -1.2,table-space-text-post = {*}]
                 S[table-format = -1.2,table-space-text-post = {*}]
                 S[table-format = -1.2,table-space-text-post = {*}]}
				\multicolumn{3}{c}{\textsc{Pearson Correlation}} \\
				\hline \hline
				{Metric} & {X1} & {X2} \\
				\hline
				\hline
				{Carroll Depression Scale} & -.321$^{**}$ & .155 \\
				\hline
				{Beck Anxiety Inventory} & -.315$^{***}$ & .389$^{***}$ \\
				\hline
				{McGill Pain Scale} & -.315$^{**}$ & .284$^{**}$ \\
				\hline
				{IBS-QOL} & -.279$^{**}$ & .151 \\
				\hline
				{IBS-SSS} & -.235$^{*}$ & .298$^{**}$ \\
				\hline
				\hline
			\end{tabular}
		}
	\end{center}
	\caption{Correlation between latent variables and other metrics taken at study baseline. $^{*}$p$<$0.05; $^{**}$p$<$ 0.01, $^{***}$p$<$0.001. 
	}
	\label{table:correlation}
\end{table}
\normalsize

\subsection{Comparison of Outcome Measures}
Given the result demonstrated in the previous section, we additionally explore the relationship between the change in latent variables and treatment impact. 

We fit linear regression models using the change in each latent variable to predict the change in each of the metrics described above. We find statistically significant results for both Beck Anxiety Inventory and IBS-SSS. Change in $\mathbf{x_1}$ could predict change in Beck Anxiety Inventory, indicating that measures to reduce symptoms such as insomnia, dry mouth, and fatigue may reduce patient anxiety. Additionally, change in $\mathbf{x_2}$ could predict change in IBS-SSS, which indicates that limiting diarrhea, loss of appetite, and headache, might have a positive impact in IBS-specific symptoms, including abdominal pain and distortion.

\begin{figure}
    \centering
    \includegraphics[width=\columnwidth]{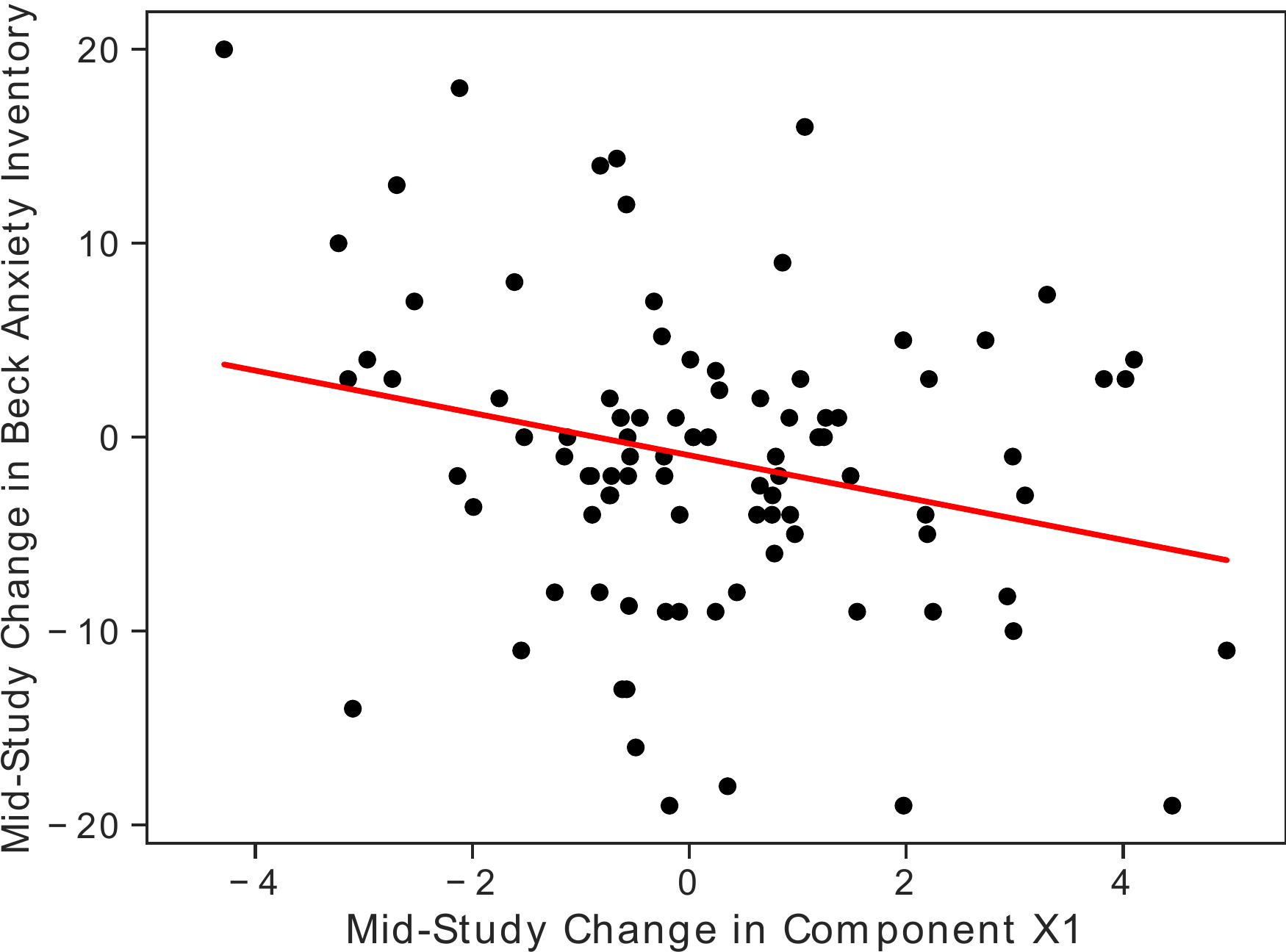}
    \caption{Linear Regression analysis of delta ${x_1}$ vs. delta Beck Anxiety Inventory. Mid-study change corresponds to values reported 3 weeks after treatment compared to baseline. 
    \newline $\Delta$ Anxiety = -1.092 ($\Delta{} x_1$) - 0.931
    \newline P $>$ $\mid t \mid$ = 0.0223;  $R^{2}$ = 0.064 }
    \label{fig:regression-anxiety}
    \vspace{-0.3cm}
\end{figure}

\begin{figure}
    \centering
    \includegraphics[width=\columnwidth]{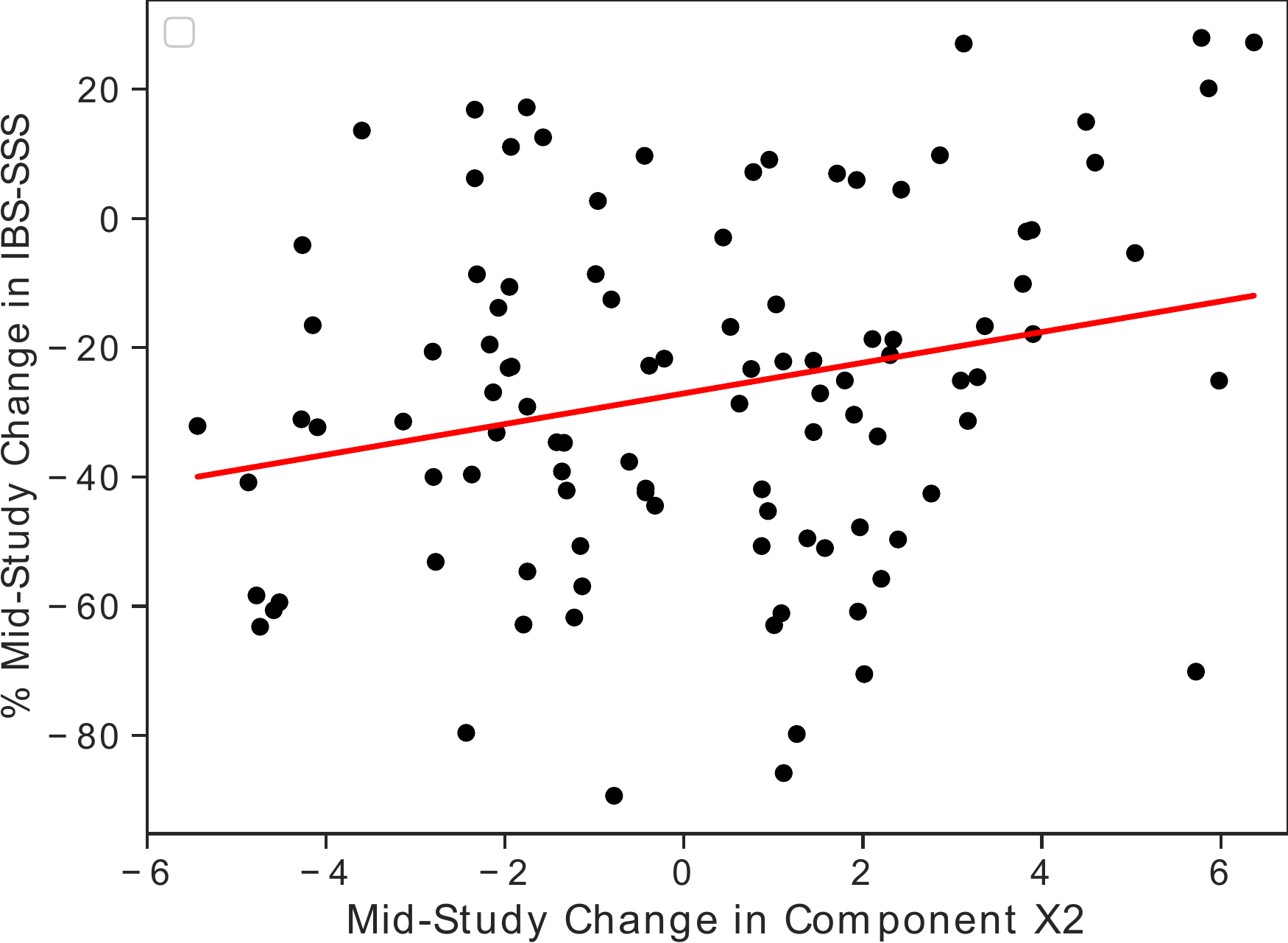}
    \caption{Linear Regression analysis of delta $x_2$ vs. delta (\%) IBS Symptom Severity Scale. Mid-study change corresponds to values reported 3 weeks after treatment compared to baseline. 
    \newline $\Delta$ IBS-SSS = 2.373 ($\Delta{} x_2$) - 27.069
    \newline P $>$ $\mid t \mid$ = 0.0148;  $R^{2}$ = 0.060
    }
    \label{fig:regression-whorwell}
    \vspace{-0.3cm}
\end{figure}

\section{CONCLUSION}
\label{section:conclusion}
We approached the challenges presented by questionnaire data through a probabilistic model framework. The output of our model is an approximately orthonormal basis, which we show is effective in visualizing groupings of responses as reported in questionnaire data, as well as latent data points which we express as an interpretable treatment outcome measure.

Through our study in IBS, we found interpretable clusters of patient symptoms, with one grouping of physical pain symptoms and another of psychological and sleep-related symptoms. Using the latent space projection for each patient resulted in statistically significant findings for both correlation with other baseline clinometrics as well as Linear Regression modeling of metric deltas.

This framework can be applied to other clinical questionnaire datasets with missing and mixed data types. In addition to the model outputs used in the application of IBS, our model may be used to obtain probability estimates for each missing data point, and the distribution of each patient data projection can also be studied to further understand the significance of these summary statistics. With each path of analysis, our solution offers interpretable results, a key element for clinical analysis which is missing from other widely used techniques such as t-SNE and UMAP.   

\addtolength{\textheight}{-12cm}

\bibliographystyle{ieeetr.bst}
\bibliography{egbib}

\end{document}